\documentclass[letterpaper, 10 pt, conference]{ieeeconf}  % Comment this line out if you need a4paper
\usepackage{soul}
\IEEEoverridecommandlockouts                              %for thanks command
\usepackage{verbatim}
\usepackage{graphics} % for pdf, bitmapped graphics files
\usepackage{amsmath,amssymb,stmaryrd,mathtools, amsfonts}
\usepackage{multirow}% http://ctan.org/pkg/multirow
\usepackage{hhline}
\usepackage{algorithm}
\usepackage{footmisc}
\usepackage[noend]{algpseudocode}
\usepackage{subfigure}  

\PassOptionsToPackage{hyphens}{url}\usepackage{hyperref} % break url at hyphens
\usepackage{dsfont}
\usepackage{algorithmicx}
\usepackage{multirow}
\usepackage[utf8]{inputenc}
\usepackage{import}
\usepackage{graphicx, graphics}
\usepackage{array}
\newtheorem{remark}{Remark}%[section]
\usepackage{color}
\usepackage[table]{xcolor}
\usepackage{siunitx}
\usepackage[short]{optidef}
\usepackage{float}
\definecolor{myedit}{rgb}{1.0,0,0}
\definecolor{mytodo}{rgb}{0,1.0,1.0}
\definecolor{lightgray}{rgb}{0.8, 0.8, 0.8}
\newtheorem{assumption}{Assumption}

\DeclareMathOperator*{\argmin}{argmin}

\newcommand{\mobility}[0]{ ${e}_{mean}$ ($\text{m/s}$)
                        & ${e}_{max}$ ($\text{m/s}$)     
                    }
\newcommand{\safety}[0]
                    { ${\mathcal{S}}_{min}$ ($\text{m}$)
                    &  ${\mathcal{P}}_{safe}$  ($\text{\%}$)}
\newcommand{\efficiency}[0]{ ${\mathcal{A}}_{mean}$ ($\text{m}/\text{s}^2$)  
                        & ${\mathcal{T}}_{solve}$ ($\text{ms}$)
                        & ${\mathcal{L}}_{long}$  ($\text{m}$)
                        }
\newcommand{\consistency}[0]{${\mathcal{P}}_{\text{LC}}$  ($\text{\%}$)}

\title{\LARGE \bf 
 Real-Time Parallel Trajectory Optimization with Spatiotemporal Safety Constraints for Autonomous Driving in Congested Traffic 
}

\author{Lei Zheng, Rui Yang, Zengqi Peng, Haichao Liu, Michael Yu Wang, \textit{Fellow, IEEE,} and Jun Ma 
    \thanks{This work was supported by the Project of Hetao Shenzhen-Hong Kong Science and Technology Innovation Cooperation Zone under Grant HZQB-KCZYB-2020083.}
    \thanks{Lei Zheng, Rui Yang, Zengqi Peng, and Haichao Liu are with the Robotics and Autonomous Systems Thrust, The Hong Kong University of Science and Technology (Guangzhou), Guangzhou, China.}
    \thanks{Michael Yu Wang is with the Department of Mechanical and Aerospace Engineering, Monash University, Clayton, Australia. }
    \thanks{Jun Ma is with the Robotics and Autonomous Systems Thrust, The Hong Kong University of Science and Technology (Guangzhou), Guangzhou, China, also with the Division of Emerging Interdisciplinary Areas, The Hong Kong University of Science and Technology, Hong Kong SAR, China, and also with the HKUST Shenzhen-Hong Kong Collaborative Innovation Research Institute, Futian, Shenzhen, China.}
    \thanks{All correspondence should be sent to Jun Ma (e-mail: jun.ma@ust.hk).}
}

\begin{document}

\maketitle
\thispagestyle{empty}
\pagestyle{empty}

%%%%%%%%%%%%%%%%%%%%%%%%%%%%%%%%%%%%%%%%%%%%%%%%%%%%%%%%%%%%%%%%%%%%%%%%%%%%%%%%
\begin{abstract}
Multi-modal behaviors exhibited by surrounding vehicles (SVs) can typically lead to traffic congestion and reduce the travel efficiency of autonomous vehicles (AVs) in dense traffic.
This paper proposes a real-time parallel trajectory optimization method for the AV to achieve high travel efficiency in dynamic and congested environments. 
A spatiotemporal safety module is developed to facilitate the safe interaction between the AV and SVs in the presence of trajectory prediction errors resulting from the multi-modal behaviors of the SVs. By leveraging multiple shooting and constraint transcription, we transform the trajectory optimization problem into a nonlinear programming problem, which allows for the use of optimization solvers and parallel computing techniques to generate multiple feasible trajectories in parallel. Subsequently, these spatiotemporal trajectories are fed into a multi-objective evaluation module considering both safety and efficiency objectives, such that the optimal feasible trajectory corresponding to the optimal target lane can be selected.
The proposed framework is validated through simulations in a dense and congested driving scenario with multiple uncertain SVs. The results demonstrate that our method enables the AV to safely navigate through a dense and congested traffic scenario while achieving high travel efficiency and task accuracy in real time.
\end{abstract}

%%%%%%%%%%%%%%%%%%%%%%%%%%%%%%%%%%%%%%%%%%%%%%%%%%%%%%%%%%%%%%%%%%%%%%%%%%%%%%%%
\section{Introduction}
\label{sec:introduction}
Autonomous vehicles (AVs) have the potential to significantly enhance driving efficiency. 
For instance, a study by the Texas A$\&$M Transportation Institute found that
the average commuters in the United States spent 54 hours stuck in traffic, costing them $\$1,008$ in wasted time and fuel in 2019~\cite{schrank2019urban}. AVs could reduce these costs and improve the overall efficiency of the transportation system. 
Decades of research and industry development in the field of autonomous driving have led to the applications of AVs in various aspects of our lives, including urban transportation~\cite{narayanan2020shared}, delivery services~\cite{liu2021role}, and mining~\cite{tian2021trajectory}, etc.  
Despite the advancements in autonomous driving technologies, 
safety remains a critical concern, particularly in dynamic urban traffic environments~\cite{shalev2017formal}. One of the key underlying factors to this concern is the multi-modal behaviors of surrounding vehicles (SVs), such as sudden deceleration and lane changes. These behaviors are difficult to predict, which significantly influence the motion of ego vehicles (EVs) and pose a safety threat to the EV.
Besides, the multi-modal behaviors of SVs can lead to traffic congestion in dense traffic scenarios, which compromises the driving efficiency of the EV.
  
\begin{figure}[tb]
    \centering
\includegraphics[width=0.95\columnwidth]{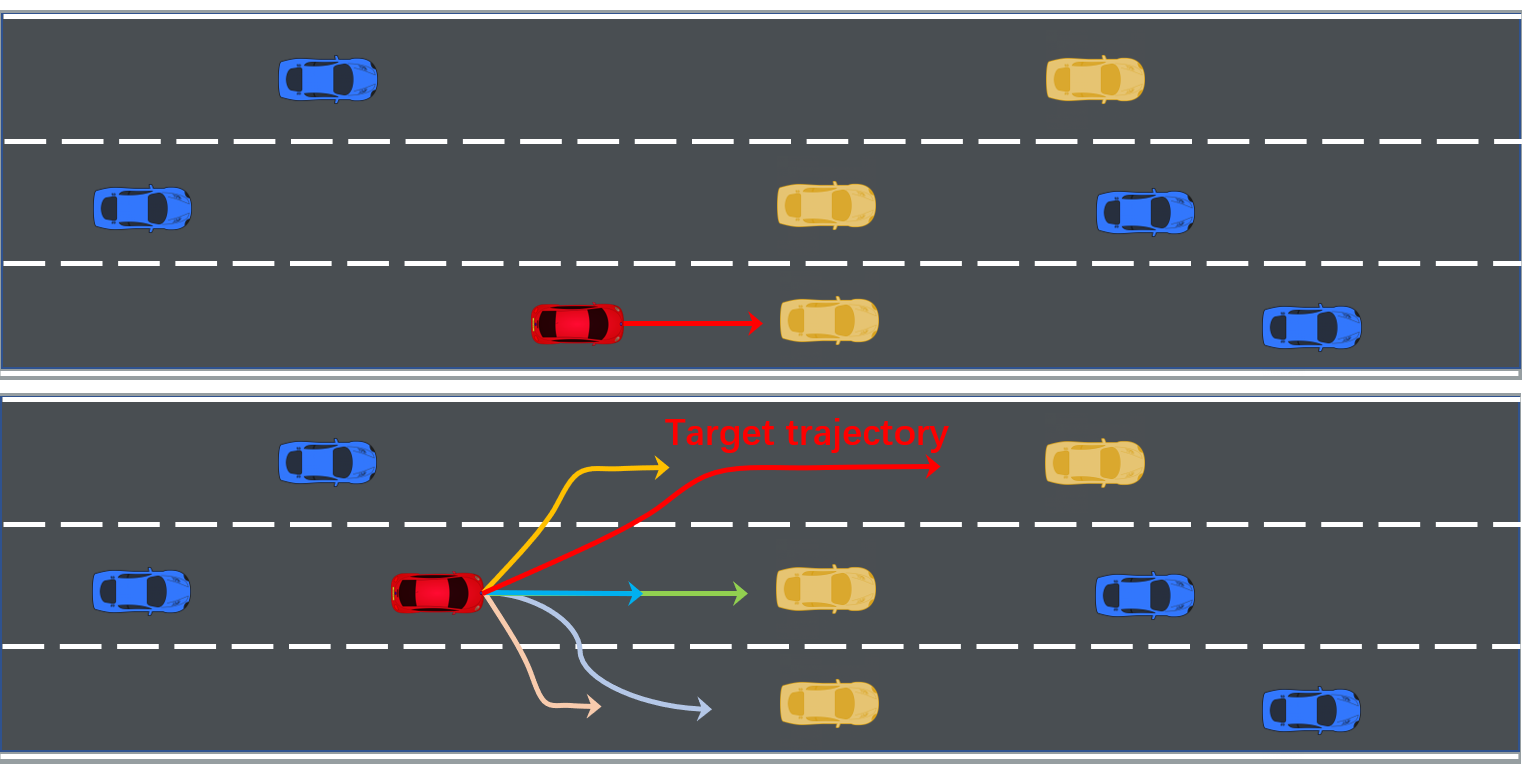}
    \caption{\small{Depiction of a three-lane driving scenario showing the red EV, yellow perceived SVs, and unperceived blue SVs. In the top subfigure, the EV is impeded by its front vehicles, while the bottom subfigure shows that with multiple parallel optimized trajectories, the EV can proactively change lanes to escape the congested scenario.}}
    \label{fig:congested_traffic} \vspace{-4mm}
\end{figure}

To overcome these challenges, the EV must be capable of rapidly replanning in response to the uncertain motion of SVs, ensuring split-second reaction to potential threats~\cite{leung2020infusing}. This requires computationally efficient trajectory optimization methods that account for the motion uncertainties of SVs. Additionally, the EV must proactively change lanes to avoid being impeded and reduce travel time, thereby improving driving efficiency, as illustrated in Fig.~\ref{fig:congested_traffic}.
  
To enable efficient replanning in traffic flow, several works have focused on decoupling longitudinal and lateral motion to generate a smooth and feasible trajectory for the EV~\cite{werling2012optimal,zhang2020optimal, sharath2020enhanced}.  
Although these works transform the motion of the EV into two independent one-dimensional movements to reduce planning complexity, this decoupling structure may result in a lack of coordination between the longitudinal and lateral movements of the EV, potentially leading to unsafe driving behaviors in sudden lane changes or abrupt braking~\cite{miller2018efficient}.
To address this issue, researchers have developed hierarchical path-speed decomposition motion planning methods that enable EVs to interact safely with other vehicles in dynamic environments~\cite{liu2017speed, fan2018baidu, jian2020multi}. These methods involve planning a path to avoid static obstacles and optimizing a speed profile to avoid dynamic obstacles~\cite{xu2021autonomous}. For example, in~\cite{jian2020multi}, an optimal path is selected among multiple candidate polynomial paths obtained from state sampling methods, and an optimal acceleration profile is generated to obtain a velocity curve based on the responsibility sensitive safety model~\cite{Gamann2019TowardsSO}. However, the path generated by the sampling module may not accurately reflect spatiotemporal information as it is not optimized for spatiotemporal safety constraints. 
This can impact the quality of the solution for the subsequent speed optimization module.
Additionally, these methods cannot fully exploit the actuator potential of the nonlinear EV with nonholonomic constraints, resulting in suboptimal control policies.

On the other hand, optimal control methods, such as Model Predictive Control (MPC)~\cite{mayne2014model}, have the ability to incorporate system models and constraints while anticipating future states for autonomous driving \cite{ zeng2021safety,zheng2022safe,yin2022trajectory}. In \cite{wurts2018collision}, the safety term is encoded into the objective function of MPC as a distance term to obstacles to avoid a potential collision and realize aggressive lane change maneuvers. To proactively avoid collisions, the control barrier function \cite{ames2019control} has been utilized to enforce the safety of the EV when overtaking the front vehicle in the MPC framework \cite{zeng2021safety}. Despite the advantages of optimal control methods in handling constraints, the computationally burdensome optimization process, particularly the inversion of the Hessian matrix, remains a limiting factor for real-time planning within typical planning horizons of 5 to 10 seconds in autonomous driving \cite{sadat2019jointly,ma2022local}.
To improve the computational efficiency towards trajectory optimization problems for autonomous driving, researchers have turned attention to the alternating direction method of multipliers (ADMM)~\cite{boyd2011distributed} to reduce the computational burden of optimization processes~\cite{ma2022alternating, han2023rda}. The ADMM splits the optimization problem into several subproblems that can be efficiently solved within the optimal control framework. While existing works often rely on a predefined motion model without uncertainties for SVs~\cite{ma2022alternating,han2023rda}, these approaches can lead to safety concerns in dynamic and congested traffic flow due to model mismatches and uncertain multi-modal behaviors of SVs. Considering uncertain behaviors and computation efficiency, an efficient multi-modal MPC approach based on the alternating minimization method has been developed for safe and efficient interaction with uncertain SVs for the EV~\cite{adajania2022multi}. It has shown promise in parallel optimization of multiple trajectories in a traffic flow with multi-modal behaviors exhibited by the SVs, but it does not explicitly address driving consistency when evaluating candidate trajectories. This could lead to frequent lane changes in dense and congested traffic, which may compromise driving comfort.

  \begin{figure}[tbp]
\centering
\includegraphics[width=0.95\columnwidth]{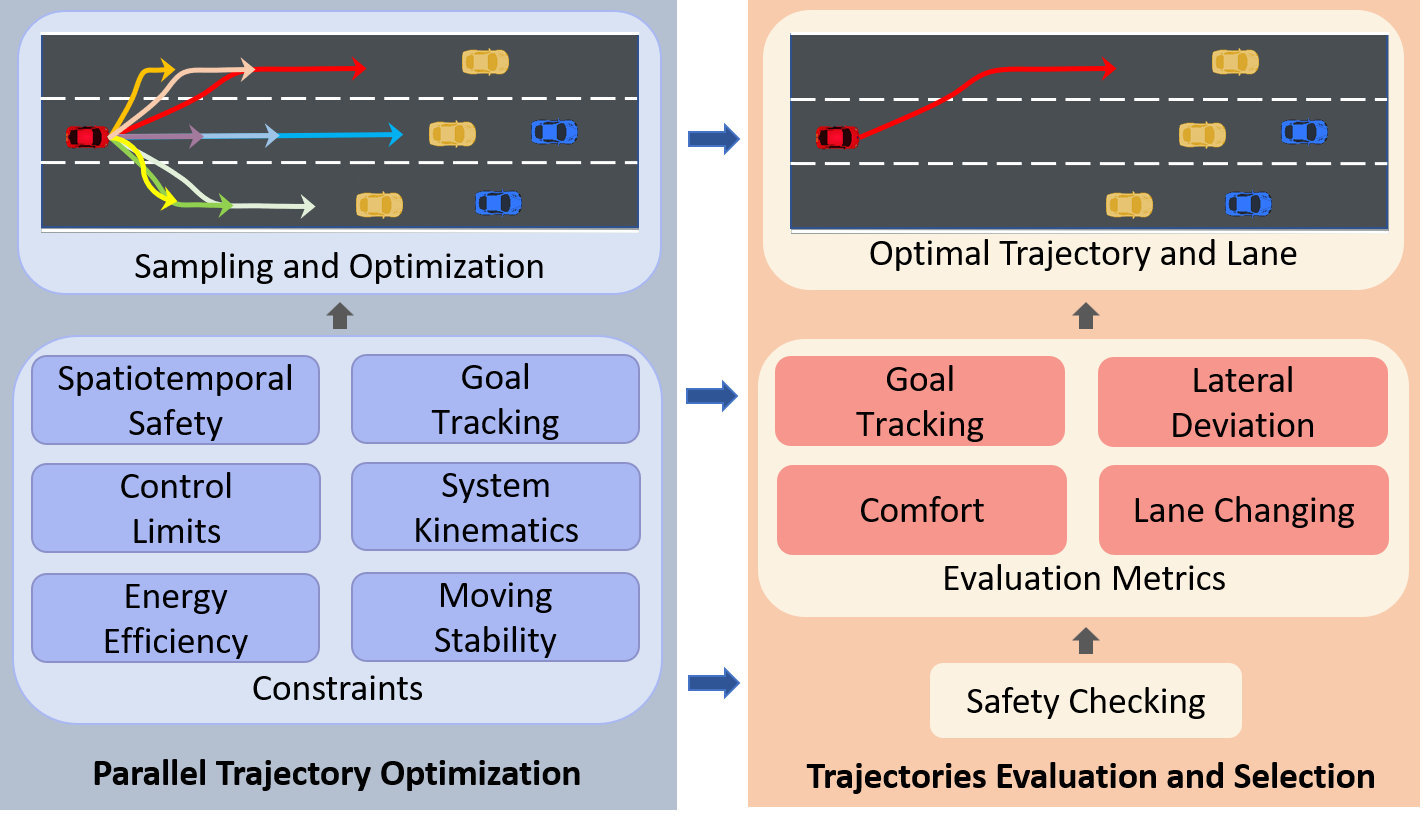}
    \caption{\small{The proposed optimization framework. The left module generates multiple candidate trajectories in parallel, with each one corresponding to a driving lane, using CPU multi-threading systems in real time. The right module evaluates the generated candidate trajectories and selects the optimal trajectory and lane based on various metrics.}} \vspace{-2mm}
\label{fig:architecture}
\end{figure}  
 
In this paper, an efficient trajectory optimization framework is proposed such that the EV is able to  navigate through dense and congested traffic safely and efficiently, as illustrated in Fig.~\ref{fig:architecture}. To facilitate efficient driving, multiple nominal trajectories are optimized simultaneously using multi-threading and direct multiple-shooting, with each trajectory corresponding to a unique target lane. Then, the optimal trajectory is evaluated and selected for the EV to execute.
 
 The main contributions of this work are as follows:
 First, we propose a computationally efficient approach that leverages spatiotemporal information between the EV and SVs and employs a multi-threading and multiple-shooting technique for parallel trajectory optimization. This allows for the generation of multiple safe and feasible energy-efficient candidate trajectories in real time while striking a balance between performance and safety.
Second, our approach prioritizes driving consistency as a crucial factor in trajectory evaluation, resulting in a smooth driving behavior that enhances driving continuity and stability, and this reduces the likelihood of abrupt lane changes. Finally, the proposed fully parallel implementation based on C++ and ROS2 achieves real-time performance in a dense and congested traffic flow. 
 
 The rest of this paper is organized as follows. The problem statement is presented in Section~\ref{sec:prblm_s}. The proposed methodology is described in Section~\ref{sec:methodology}. The numerical simulation of the proposed algorithm on an autonomous vehicle system is shown in Section~\ref{sec:simulation}. Finally, the conclusion is drawn in Section~\ref{sec:conclusion}.

% \input{Sections/related_work}
% \newpage
\section{Problem Statement}
\label{sec:prblm_s}
\begin{figure}[tbp]
\centering
\includegraphics[width=0.95\columnwidth]{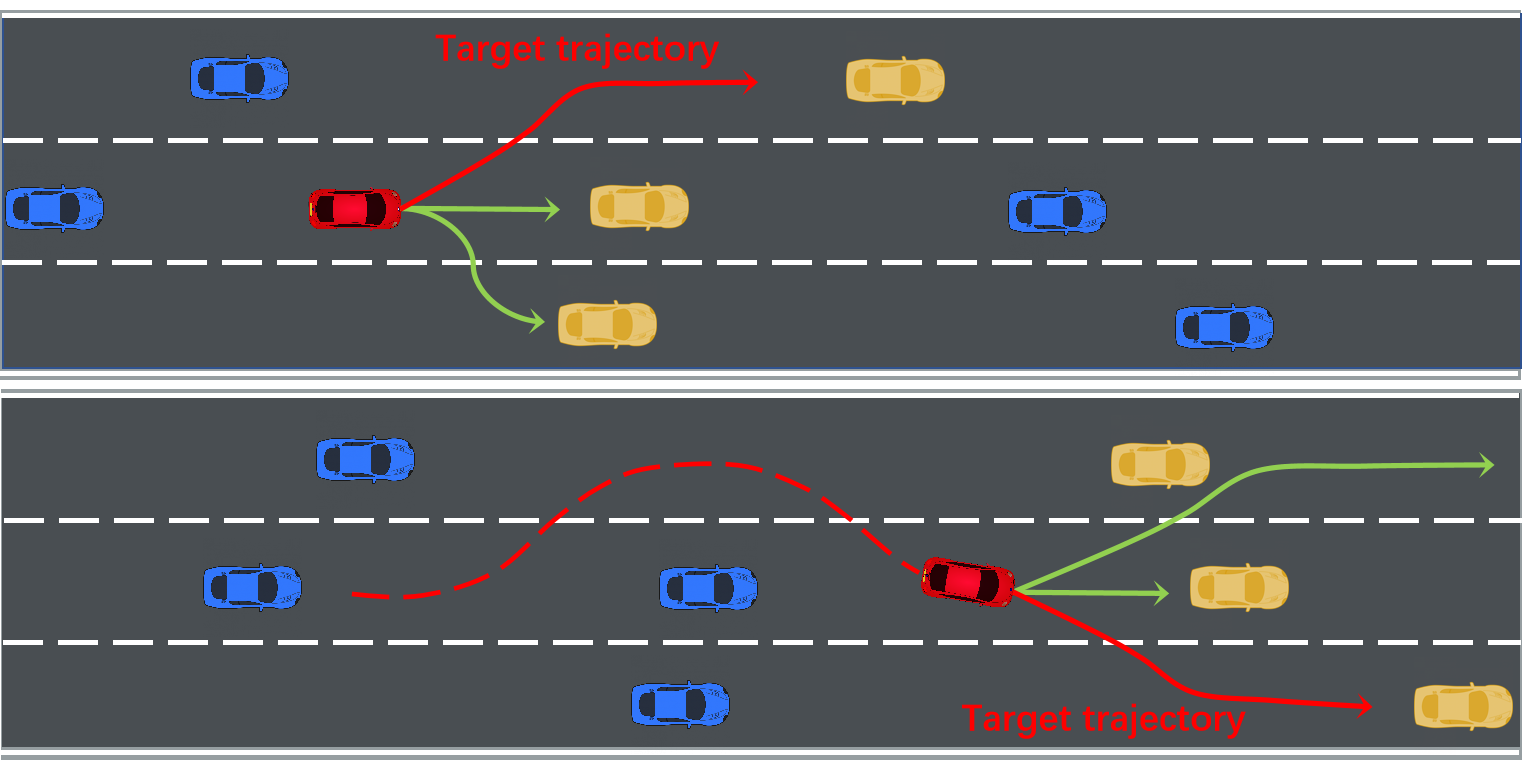}
    \caption{\small{Illustration of the motion of a red EV in a dynamic congested scenario with three lanes. Yellow and blue vehicles represent perceived and unperceived vehicles, respectively. The dashed red line shows the EV's executed trajectory, while the solid red line represents the target trajectory. Other lines with an arrow denote unselected candidate trajectories of the EV.}} \vspace{-2mm}
\label{fig:Problem_statement}
\end{figure}

In this section, we consider a three-lane dense and congested drive scenario, as illustrated in Fig.~\ref{fig:Problem_statement}. 
The EV is represented by a nonlinear kinematic bicycle model~\cite{chen2017constrained}. The state vector of the ego vehicle is defined as follows:
\begin{equation}
\textbf{x}=[p_x, p_y, \theta, v, \omega]^T \in \mathcal{X},
\label{eq:ev_state}
\end{equation}
where $p_x$ and $p_y$ denote the longitudinal and lateral positions in the global coordinate, respectively; $\theta$, $v$, and $\omega$ denote the heading angle, velocity, and yaw rate, respectively. The control input vector to the EV is defined as $\textbf{u}=[a, \dot{\omega}]^T\in \mathcal{U}$, where $\omega = \frac{v \tan(\delta)}{L}$, $L$ is the wheelbase, $a$ is the acceleration, and $\delta$ is the steering angle of the front wheels. The nonlinear kinematic bicycle model can be formulated as follows:
\begin{equation}
\Dot{\textbf{x}}=\left[\begin{array}{c}
\dot{p}_x\\
\dot{p}_y\\
\dot{\theta}\\
\dot{v}\\
\dot{\omega}
\end{array}\right]=\left[\begin{array}{c}
v\cos \left( \theta\right)\\
v\sin \left(\theta \right)\\
\omega\\
0\\
0
\end{array}\right] + \left[\begin{array}{cc}
0&0\\
0&0\\
0&0\\
1&0\\
0&1
\end{array}\right] \left[\begin{array}{c}
a \\
\dot{\omega}
\end{array}\right].
\label{eq:ev_model}
\end{equation}

In this dense and congested driving scenario, the EV may be impeded by its front vehicles with uncertain behaviors (e.g., velocity keeping, acceleration, deceleration), resulting in poor driving efficiency and compromised safety. Besides, frequent lane changes in such scenarios may also destabilize the driving experience and threaten the EV's passengers' safety.
To make this problem tractable, we make the following assumptions:
\begin{assumption}(\textbf{Safety\ Responsibility}~\cite{shalev2017formal}) 
\label{assumption: Safety_responsibility} When the two vehicles are driving in the same direction, if the rear vehicle $c_r$ hits the front vehicle $c_f$ from behind, then the rear vehicle $c_r$ is responsible for the accident.
\end{assumption} 
\begin{assumption}(\textbf{Driving \ Maneuver}~\cite{shalev2017formal})
% ~\cite{shalev2017formal})
\label{assumption: communication} 
A vehicle will respond appropriately to dangerous situations and not hit another vehicle from behind as long as the front vehicle does not brake stronger than the maximum deceleration of the vehicle behind.
\end{assumption} 
    \begin{assumption}(\textbf{Perception\ Ability})\label{assumption: communication}
An EV can obtain information about the position and speed of the three nearest SVs.
\end{assumption} 

To tackle these challenges, we aim to design an efficient trajectory planning framework that enables the EV to safely interact with SVs while keeping high travel efficiency. The framework generates optimized trajectories in parallel, which closely follow the centerline of different target lanes. Each trajectory $\tau (t)\in \mathcal{T}$ corresponds to a specific target lane $\xi \in \Xi$, given as follows:
\begin{equation}\label{eq:nominal_trajectories}
 \mathcal{T}:= \left\{\textbf{u}^{(j)}(t), \textbf{x}^{(j)}(t)\right\}_{j=1}^{N_t},\ 0\leq t\leq T,
\end{equation}\vspace{-4mm}
\begin{equation}\label{eq:target_lanes}
	\Xi:= \{lane1, \ lane2,\ lane3\},
\end{equation}
where $T$ and $N_t$ represent the optimization horizon and the number of trajectories, respectively; $ \Xi$ denote the set of driving lanes, where each target lane represents either a lane-keeping maneuver, a left-lane-change, or a right-lane-change maneuver for the EV, as depicted in Fig.~\ref{fig:Problem_statement}.
 
The proposed parallel trajectory optimization framework operates over a planning horizon $T$, generating trajectory $\tau$ for each target lane $\xi \in \Xi$. The goal is to determine the optimal lane $\xi^*$ and trajectory $\tau^*$ in real time to achieve the following objectives: 
 \begin{itemize} 
    \item [(O1)] 
 \emph{Efficiency:} The proposed framework can parallel optimize multiple reference trajectories along the center lane in real time, enabling the EV to navigate through congested traffic with high travel efficiency. 
    \item [(O2)] 
\emph{Safety:} The proposed strategy can leverage spatiotemporal information to ensure safe navigation through dynamic congested traffics while maintaining a safe distance from other vehicles.
    \item [(O3)] \emph{Consistency and Stability:} The proposed approach should ensure spatiotemporal consistency to avoid frequent lane changes, reducing the likelihood of abrupt maneuvers that may affect driving stability. 
\end{itemize}

\section{Methodology}
\label{sec:methodology}
% We propose a parallel trajectory optimization framework to achieve the three goals in Section~\ref{sec:prblm_s}, as illustrated in Fig.~\ref{fig:architecture}. 
In this section, we first describe the goal-oriented movement of the EV in Section~\ref{subsec:lane_oriented_module}, which involves designing appropriate costs to achieve desired tasks. Following up on our previous work~\cite{zheng2023STRHC}, we introduce the spatiotemporal safety module for collisions avoidance in Section~\ref{subsec:safety_module}. Subsequently, we present the parallel sampling and trajectory optimization method in Section~\ref{subsec:parallel_opt_module}. Lastly, the decision-making module that aims to select the optimal trajectory and the target lane is presented in Section~\ref{subsec:decision_making}.

\subsection{Goal-Oriented Movement}\label{subsec:lane_oriented_module}
The goal-oriented movement module is responsible for generating trajectories aligned with the target lanes' centerline. To achieve this, we sample several points along each target lane at the end of the planning horizon and use them as constraints to generate candidate trajectories that align with the target lanes' centerline, as illustrated by the arrow of each trajectory in the left module of Fig.~\ref{fig:architecture}. 

To ensure that each optimized trajectory remains close to the centerline of the lane, we enforce the final lateral position of each trajectory to match the lateral position of corresponding sampling points. This is accomplished by setting the terminal cost to ensure stability in the quasi-infinite horizon approach \cite{de1998stabilizing}. Likewise, we can set the EV's final target yaw angle and yaw rate state to tiny values to endow the EV with a stable driving mode. Therefore, we introduce a terminal cost to enforce the final state of each trajectory in the vicinity of the desired state as follows: 
\begin{eqnarray}
\left\{
    \begin{aligned}
        C_T &= \phi(\textbf{x}(t+T)) = (\iota_T \textbf{x}(t+T))^T \textbf{Q}_T (\iota_T \textbf{x}(t+T)), \\
        C_T &> \int_{0}^{T} \mathcal{L}(\textbf{x}(t),\textbf{u}(t)) dt,  
    \end{aligned}
\right.
\label{eq:terminal_cost}
\end{eqnarray}
where $\textbf{x}(t+T)$ is the terminal state vector; $\textbf{Q}_T$ is a weighting matrix; $\iota_T = diag(0,1,1,0,1)$ extracts the lateral position, heading angle, and yaw rate from the state vector; $\mathcal{L}(\textbf{x}(t),\textbf{u}(t))$ represents the running cost at time $t$. 
% The primary idea is to choose the terminal cost to be larger than the running cost~\cite{fontes2001general}. Hence we set 
$\textbf{Q}_T$ is set to be a large value to ensure the terminal cost is larger than the running cost~\cite{fontes2001general}. 
 
To further enhance driving task performance, we introduce a goal-tracking cost term for each trajectory
% that incorporates the desired state $\textbf{x}_d$ of the EV 
as follows:
\begin{equation}
     C_m(t) = (\iota_m(\textbf{x}(t)-\textbf{x}_d(t)))^T\textbf{Q}_m \iota_m(\textbf{x}(t)-\textbf{x}_d(t)),
\label{cost_cruise}
\end{equation}
where $\textbf{x}_d(t)$ denotes the desired state vector at time $t$ that the EV aims to reach to accomplish its task; ${\iota}_m$ extracts the particular state vector of the EV from the state vector $\textbf{x}$. 
For instance, in a cruise task, we set $\iota_m=diag(0,1,0,1,0)$ to extract the lateral position and velocity from the state vector $\textbf{x}$. This enables the EV to cruise at a desired speed and drive close to the centerline of the lane.  
\subsection{Spatiotemporal Safety Module}\label{subsec:safety_module} 
To realize safe interaction between the EV and the $i$th SV, we need to keep the EV's state within a safe region $\mathcal{S}$ defined as follows:
\begin{equation}
    \mathcal{S}:=\{\textbf{x}(t)\in\mathcal{X}|h(\textbf{x}(t), \textbf{O}_i(t))\geq0\},
\end{equation}
where $\textbf{O}_i={[{\textbf{O}_{p,i}}, {\textbf{O}_{v,i}}]}^T $; $\textbf{O}_{p,i} = [o_{x,i}, o_{y,i}] $ and $\textbf{O}_{v,i} = [o_{v_{x},i}, o_{v_{y},i}]$ denote the position and velocity vectors of the $i$th SV, which evolves according to an uncertain system model $f^{SV}(\cdot)$; $t$ denotes current time; $h:\mathbb{R}^{n}\rightarrow\mathbb{R}$, is a continuously differentiable safety barrier function that encodes state constraints as follows:
\begin{equation}
    h(\textbf{x}(t), \textbf{O}_i(t)) = 
    \frac{(p_x(t) - o_{x,i})^2}{a^2}+\frac{(p_y(t) - o_{y,i})^2}{b^2}-1,
    \label{eq:barrier_function}
\end{equation} 
 where $a$ and $b$ represent the major and minor axis lengths of a safe ellipse encompassing the form of the EV, respectively.
 
Since the laws of motion govern the SVs' positions, we heuristically infer that their nominal trajectory over the planning horizon is based on a constant velocity model in a planning horizon as follows:
\begin{equation}
   \Bar{f}^{SV}(\textbf{O}_i(t)) =\left[\begin{array}{c}
\dot{o}_{x,i}\\
\dot{o}_{y,i}
\end{array}\right]=\left[\begin{array}{c}
\dot{o}_{v_x,i}\\
\dot{o}_{v_y,i}
\end{array}\right].
    \label{eq:nominal_sv_model}
\end{equation} 
   
Nevertheless, the multi-modal uncertain behaviors of SV are hard to model accurately, especially with a simple constant velocity model, which may lead to more significant errors as the prediction horizon becomes longer. 
In this context, considering safety equally in each time step during the planning horizon can lead to overly conservative actions that compromise driving efficiency. 
To compensate for these errors and strike a balance between safety and performance in trajectory planning, we introduce a time-varying discount weight $w_i(t)$ in the safety constraint, given by: 
\begin{equation}
w_i(t) = \lambda_i \exp\left(\frac{-t}{\gamma}\right),
    \label{eq:discount_weight}
\end{equation}  
where $\gamma$ is a constant discount factor, and $\lambda_i$ is a factor representing the safety weight regarding the $i$th SV. 

With this discount weight $w_i(t)$ and the barrier function~(\ref{eq:barrier_function}), we introduce a spatiotemporal safety cost $C_{s}(t)$ as follows:
\begin{equation}
    C_{s}(t) = \sum_{i=1}^{M} w_i(t)H(\textbf{x}(t), \textbf{O}_i(t)),
    \label{eq:safety_function}
\end{equation}  
 where $M$ denotes the number of the perceived SVs, and $H(\textbf{x}(t), \textbf{O}_i(t))$ is a safety measurement function based on the barrier function~(\ref{eq:barrier_function}), as follows:
\begin{equation}
        \small
    \label{eq:safety_function}
    \begin{split}
    H(t) = \\ 
     &\frac{1}{\eta + h(\textbf{x}(t), \textbf{O}_i(t))}\left(1 - \frac{h(\textbf{x}(t), \textbf{O}_i(t)) - c}{\varepsilon + \sqrt{(h(\textbf{x}(t), \textbf{O}_i(t))- c)^2}}\right),
    \end{split}
    \end{equation}
 where $\eta \in\mathbb{R^{+}}$ denotes a scale factor, and $\varepsilon \in\mathbb{R^{+}}$ is a small regularization constant facilitating numerical stability.
\begin{remark}
    Although the spatiotemporal cost terms $C_s$ are technically soft, they act like hard constraints as the penalty is obtained immediately after the constraint boundary is violated. However, unlike hard constraints, they have the advantage that the importance of different safety requirements regarding different time steps in a horizon can be adjusted by setting different weighting matrices  $w_i(t)$.
\end{remark}  
\subsection{Parallel Trajectory Optimization Framework}\label{subsec:parallel_opt_module} 
We design an optimal control framework to model the parallel trajectory optimization problem. Then, we transform the parallel optimal control problem into a parallel nonlinear programming (NLP) problem to optimize the nominal sampled trajectories based on the direct multiple shooting method~\cite{ betts2010practical}. 
Specifically, the optimal control for the $j$th trajectory takes the following form:
 \begin{mini!}[2]
 {\substack{  \textbf{u}^{(j)}(t), \textbf{x}^{(j)}(t) 
 }}{\int_{0}^{T} \mathcal{L}^{(j)}(\textbf{x}(t),\textbf{u}(t))dt+ \phi^{(j)}(\textbf{x}(T))\label{opt:obj}}{\label{opt:SMPC_skeleton}}{}
 \addConstraint{\textbf{x}(0)=\textbf{x}_0,\ \textbf{O}_i(0)=\textbf{O}_{i,0}}{\label{opt:init}}{}
\addConstraint{\dot{\textbf{x}}(t)}{= f^{EV}(\textbf{x},\textbf{u})\label{opt:EV_dyn}}
\addConstraint{\dot{\textbf{O}}_i(t)}{= \Bar{f}^{SV}(\textbf{O}_i)\label{opt:SV_dyn}}
\addConstraint{\textbf{x}_{min} \preceq  \textbf{x}^{(j)}(t) \preceq \textbf{x}_{max} }{\label{opt:state_limit}}
\addConstraint{\textbf{u}_{min} \preceq  \textbf{u}^{(j)}(t) \preceq \textbf{u}_{max}}{\label{opt:control_limit}}
\addConstraint{(\textbf{u}^{(j)}(t), \textbf{x}^{(j)}(t) )\in\mathcal{U}\times\mathcal{X}}{\label{opt:domain}}
\addConstraint{~ \forall t\in [0, T],}{\nonumber}
\end{mini!}
where $\textbf{x}_0$ and $\textbf{O}_{i,0}$ denote the initial state of the EV and the $i$th SV, respectively; (\ref{opt:EV_dyn}) is the nonlinear motion model of the EV defined in (\ref{eq:ev_model}); $\textbf{x}_{min}$ and $\textbf{x}_{max}$ denote the minimum and maximum state constraints of the EV, respectively; $\textbf{u}_{min}$ and $\textbf{u}_{max}$ denote the minimum and maximum control inputs of the EV, respectively; $\phi^{(j)}(\textbf{x}(T)) = C^{(j)}_T$ represents the terminal cost for the $j$th trajectory.  
The running cost of the $j$th trajectory at time $t$ is denoted by $\mathcal{L}^{(j)}$ in the form: 
\begin{equation}
\mathcal{L}^{(j)}(\textbf{x}(t),\textbf{u}(t)) = C^{(j)}_m(t)  +   C^{(j)}_s(t) + \|\textbf{u}^{(j)}(t)\|^2_{\textbf{R}},
    \label{eq:running_cost_function}
\end{equation}  
where $C^{(j)}_m(t)$, $C^{(j)}_s(t)$, and $\|\textbf{u}^{(j)}(t)\|^2_{\textbf{R}}$ denote the goal-tracking cost, spatiotemporal safety cost, and energy consumption cost for the $j$th trajectory, respectively; $\textbf{R}$ is a positive semi-definite diagonal matrix determining energy efficiency.

We further reformulate the initial optimal control
problem (\ref{opt:obj})-(\ref{opt:domain}) into a multiple shooting-based constrained NLP problem to efficiently and accurately handle nonlinear
EV's kinematics~(\ref{eq:ev_model}) to achieve efficient numerical solutions for parallel trajectory optimization~\cite{borrelli2017predictive}.
The fundamental idea of direct multiple shooting is to break the trajectory into $N$ shorter segments. By doing this, the overall trajectory optimization problem can be transformed into $N$ smaller shooting intervals that span the optimization horizon $T$~\cite{betts2010practical}. 
Then, continuity constraints between two shooting intervals are enforced as follows:
\begin{equation} 
\Bar{\textbf{x}}_{k+1}: = f(\textbf{x}_k,\textbf{u}_k)\ominus
\textbf{x}_{k+1} = \textbf{0},
\end{equation}
where $\ominus$ denotes the difference operator
of the state manifold that is needed
to optimize over manifolds~\cite{mastalli2022feasibility}; $f(\textbf{x}_k,\textbf{u}_k)$ denotes the simulation of the
nonlinear kinematics~(\ref{eq:ev_model}) over one interval, starting from state $\textbf{x}_k$ with control input $\textbf{u}_k$. In this study, we utilize 4th-order Runge-Kutta integration with a
sampling time of $T_s$ = 100 $\,\text{ms}$ to discretize the original system~(\ref{eq:ev_model}).
The resulting NLP problem, which enables parallel trajectory optimization based on multi-threading computation, can be formulated as follows:
 \begin{mini!}[2]
 {\substack{u^{(j)}_k,x^{(j)}_{k+1}}}{C^{(j)}_T  + \sum_{k=0}^{N-1} \left( C^{(j)}_{m,k} + \sum_{i=1}^{M} C^{(j)}_{s,k} + \|\textbf{u}^{(j)}_k\|^2_{\textbf{R}} \right) \label{eq:obj}}{\label{opt:PTO}}{}
  \addConstraint{\textbf{x}(0)=\textbf{x}_0,\ \textbf{O}_i(0)=\textbf{O}_{i,0}}{\label{eq:init}}{}
\addConstraint{\Bar{\textbf{x}}_{k+1}: = f(\textbf{x}_k,\textbf{u}_k)\ominus
\textbf{x}_{k+1} = \textbf{0}\label{eq:EV_dyn}} 
\addConstraint{O_{k+1}}{= \hat{f}^{SV}(\textbf{O}_{k})\label{eq:SV_dyn}}
\addConstraint{\textbf{x}_{min} \preceq  \textbf{x}^{(j)}_k\preceq \textbf{x}_{max} }{\label{opt:state_limit}}
\addConstraint{\textbf{u}_{min} \preceq  \textbf{u}^{(j)}_k \preceq \textbf{u}_{max}}{\label{eq:control_limit}}
\addConstraint{(\textbf{u}^{(j)}_{k},\textbf{x}^{(j)}_{k+1})\in\mathcal{U}\times\mathcal{X}}{\label{eq:domain}}{}
\addConstraint{~ \forall k\in\mathcal{I}_0^{N-1},}{\nonumber}
\end{mini!}
where $\hat{f}^{SV}$ is a discrete form of the predictive model of SVs (\ref{eq:nominal_sv_model}) based on an Euler method.

We can optimize the NLP problem (\ref{eq:obj})-(\ref{eq:domain}) for each nominal trajectory in a parallel manner using multi-threading computation. Hence, we can obtain optimized candidate trajectories $\left\{\tau_j\right\}_{j=1}^{N_t}$, where $\tau_j = \left\{\textbf{u}^{(j)}_k, \textbf{x}^{(j)}_{k+1}\right\}_{k=0}^{N-1}$; each trajectory $\tau_j$ corresponds to a target lane $\xi_j$ enforced by the terminal constraint $C^{(j)}_T$.
  
To achieve efficient trajectory optimization to realize fast replanning that enables the EV to respond quickly and robustly to dynamic environments and further mitigate the effects of trajectory prediction errors of the SVs, we implement a receding horizon approach to optimize the NLP problem (\ref{eq:obj})-(\ref{eq:domain}) based on sequential quadratic programming (SQP). 

% At each time step, the EV will execute only the first control input $\textbf{u}^{*}_0$ in the optimized sequence $ \left\{\textbf{u}^{*}_k\right\}_{k=0}^{N-1}$ obtained from $N_t$ evaluated candidate trajectories. The evaluation will be discussed in subsequent Section~\ref{subsec:decision_making}.
At each time step, the EV will execute only the first control input $\textbf{u}^{*}_0$ in the optimized sequence $ \left\{\textbf{u}^{*}_k\right\}_{k=0}^{N-1}$, which is obtained from the optimal trajectory among $N_t$ candidate trajectories evaluated in the subsequent Section~\ref{subsec:decision_making}. 
The remaining control sequence $ \left\{\textbf{u}^{*}_k\right\}_{k=1}^{N-1}$ is subsequently utilized as an initial solution for the warm start of the NLP problem (\ref{eq:obj})-(\ref{eq:domain}), thereby enhancing the convergence speed.
\begin{remark}
The quadratic structure of the objective function (\ref{eq:obj}) allows for efficient optimization using the Hessian matrix information. However, computing the exact inverse Hessian matrix is often computationally burdensome in real-time applications. Instead, we utilize Gauss-Newton methods~\cite{gratton2007approximate} to approximate the inverse Hessian matrix from gradient information, facilitating more efficient optimization.
\end{remark} 
  
\subsection{Decision-Making with Consistency}
\label{subsec:decision_making}
After checking the safety of the next position of each candidate trajectory based on the barrier function (\ref{eq:barrier_function}), we aim to determine the optimal decision behavior $\xi^*$ and trajectory $\tau^*$ for the EV. Specifically, we design an evaluation algorithm that considers driving efficiency and stability for the optimized candidate trajectories obtained in (\ref{eq:obj})-(\ref{eq:domain}) as follows:
\begin{equation}\label{eq:optimal_behavior}
	\begin{aligned}
    \{ \xi^*, \tau^*\}	= \argmin_{\substack{\xi_j \in \Xi,\tau_j \in \mathcal{T}}}&\  \displaystyle s(\xi_j, \tau_j, \textbf{w}),  
	\end{aligned}
\end{equation}
where the overall cost function $s(\xi, \tau,\textbf{w})$ is defined as: 
\begin{equation}\label{eq:optimal_cost_func}
	\begin{aligned}
    s(\xi_j, \tau_j,\textbf{w})= \textbf{w}^T \textbf{f}(\xi_j, \tau_j),
	\end{aligned}
\end{equation}
where the weight vector $\textbf{w} = [w_g, w_l, w_c, w_m]^T$ determines the relative importance of each sub-cost for a trajectory; $\textbf{f}(\xi_j, \tau_j )$ denotes a vector of sub-costs that captures various aspects of the $j$th trajectory's performance as follows:
\begin{equation}\label{eq:evaluation_func}
 \textbf{f}(\xi_j, \tau_j) = [F_g,  F_l , F_c ,  F_m]^T,
\end{equation}
 where $F_g$, $F_l$, $F_c$, and $F_m$ represent the goal-tracking, lateral deviation, comfort, and consistency costs, respectively. 
\subsubsection{Goal-tracking metric} 
The goal-tracking cost evaluates the ability of the  optimized candidate trajectories to accomplish a driving task. For a cruise driving task, this cost aims to minimize the difference between the actual velocity of the EV and the target cruise velocity $v_g$. To account for the motion uncertainty of SVs over a long horizon, we apply an exponentially decreased weight with a discount factor $\gamma_g$ for the longer $N-N_c$ time steps. Thus, we can compute an initial cost for this term using the following equation:
\begin{equation}
\small
C_g = \sum_{i=1}^{N_c-1}||v_i - v_{g}||^2 + \sum_{i=N_c}^{N}\exp\left(\frac{-(i-N_c)}{\gamma_g}\right) ||v_i - v_{g}||^2,
\end{equation}
where $v_i$ denotes the actual velocity of the EV at time step $i$; $N_c$ represents the number of initial time steps with relatively reliable position prediction of SVs at the beginning of the optimization horizon, which has a total of $N$ time steps (e.g., $N_c =10$ when $N=50$).
 
To compare the goal-tracking cost among different sub-costs, we normalize $C_g$ for each trajectory. This eliminates the influence of different units on performance metrics, giving us the goal-tracking cost $F_g$ as follows:
\begin{equation}
F_g = \frac{C_{g}- C_{g,min}}{C_{g,max}- C_{g,min}},
\end{equation}
where $C_{g,min}$ and $C_{g,max}$ are the minimum and maximum values of $C_g$ across all candidate trajectories, respectively.
  
\subsubsection{Lateral deviation metric}
The lateral deviation cost penalizes the distance of the optimized candidate trajectories to the center of each lane. The goal of this cost is to keep the EV close to the centerline, thereby enabling the EV to drive more stably and reduce the risk of collision with other vehicles. It can be represented by a function $C_l$ that depends on the lateral position of the EV relative to the centerline of each target lane as follows,
\begin{equation}
\small
C_l = \sum_{i=1}^{N_c-1}||p_{y,i} - y_{c}||^2 + \sum_{i=N_c}^{N}\exp\left(\frac{-(i-N_c)}{\gamma_l}\right) ||y_i - y_{c}||^2,
\end{equation}
where $p_{y,i}$ is the lateral position of the EV at time step $i$; $y_{c}$ is the lateral position of the centerline of the lane; $\gamma_l$ is a discount factor.
Similar to the goal-tracking cost, we normalize $C_l$ for each trajectory using the following lateral deviation cost:
\begin{equation}
F_l = \frac{C_{l}- C_{l,min}}{C_{l,max}- C_{l,min}},
\end{equation}
where $C_{l,min}$ and $C_{l,max}$ are the minimum and maximum values of $C_l$ among all candidate trajectories, respectively.

\subsubsection{Comfort metric} The comfort cost penalizes the discomfort experienced by passengers due to the change in acceleration. The goal of this cost is to minimize the acceleration change. It can be represented by a function $C_c$ that depends on the jerk of the EV as follows,
\begin{equation}
\small
C_c = \sum_{i=1}^{N_c-1}||j_i||^2 + \sum_{i=N_c}^{N}\exp\left(\frac{-(i-N_c)}{\gamma_c}\right) ||j_i||^2,
\end{equation}
where $j_i=\frac{a_{i+1}-a_i}{T_s}$ is the jerk of the EV at time step $i$; $\gamma_c$ is a discount factor.
We normalize $C_c$ for each trajectory to get the comfort cost as follows:
\begin{equation}
F_c = \frac{C_{c}- C_{c,min}}{C_{c,max}- C_{c,min}},
\end{equation}
where $C_{c,min}$ and $C_{c,max}$ are the minimum and maximum values of $C_c$ across all candidate trajectories, respectively.
\subsubsection{Consistency metric}
This cost penalizes frequent lane-changing behaviors, which may result in erratic and inconsistent driving. It can be represented by a function $C_m$ that depends on the last target lateral position $y_{g,k-1}$ of last target lane $\xi_{k-1}$ and new target lateral position $y_{g,k}$ of goal lane $\xi_{k}$ as follows: 
% \begin{eqnarray}
% \left\{
%     \begin{aligned}
%         C_m &=  ||y_{g,k} - y_{g,k-1}||^2, \\
%          y_{g,k} & = y_{c,k},  
%     \end{aligned}
% \right.
% \label{eq:lane_changing_cost}
% \end{eqnarray}
\begin{equation}
    C_m =  ||y_{g,k} - y_{g,k-1}||^2, 
    \label{eq:lane_changing_cost}
\end{equation} 
where $k$ denotes the current time step of driving.
Similarly, we normalize $C_c$ for each trajectory using the following equation:
\begin{equation}
\small
F_m = \frac{C_{m}- C_{m,min}}{C_{m,max}- C_{m,min}},
\end{equation}
where $C_{m,min}$ and $C_{m,max}$ are the minimum and maximum values of $C_m$ among all candidate trajectories, respectively.

\section{Simulation Study}
\label{sec:simulation}
In this section, we evaluate the effectiveness of our proposed parallel trajectory optimization framework in an adaptive cruise driving task under a dense and congested traffic flow.     
\subsection{Simulation Setup}
We conduct the simulation experiments using ROS2 on an Ubuntu 22.04 LTS system environment with an AMD Ryzen 5 5600G CPU with 6 cores and 12 threads running at a clock speed of  3.26 $ \,\text{GHz}$ and 16 GB of RAM.
The control and communication frequencies between the EV and SVs are set as 10$\,\text{Hz}$. The simulation time and optimization horizon $T$ are set as $20\,\text{s}$ and $5\,\text{s}$, respectively.
We utilize the state-of-the-art optimization solver Acado~\cite{Houska2011a} as the SQP solver for the NLP problem (\ref{eq:obj})-(\ref{eq:domain}) for each nominal trajectory in a parallel manner using multi-threading computation in C++. 
   
The EV aims to navigate through a dense and congested traffic flow in a one-direction three-lane road while avoiding the other nine SVs and keeping a target cruise speed $v_g = 15 \,\text{m/s}$. 
The SVs follow the intelligent driver model adopted from~\cite{adajania2022multi} and move parallel to the
centerline while adapting their cruise velocity based on
the distance to the cars in front. The initial states and target speeds of SVs are set as follows:   
\begin{itemize} 
    \item Initial States: 
    $\textbf{O}_0={[-10 , -10, 9.5,0 ]}^T$, $\textbf{O}_1={[ 25, -10,8.5,0]}^T $, $\textbf{O}_2={[ 60, -10, 9.0,0]}^T$, $\textbf{O}_3={[70 , -6, 8,0 ]}^T$, $\textbf{O}_4={[ 85, -6,8.5,0]}^T$, $\textbf{O}_5={[ 100, -6,9.2,0]}^T $, $\textbf{O}_6={[130 , -2, 10, 0]}^T$, $\textbf{O}_7={[110, -2,8,0]}^T$, $\textbf{O}_8={[ 160, -2, 12,0]}^T$.
    \item The target longitudinal speed is set as $10 \,\text{m/s}$, $8\,\text{m/s}$, $12\,\text{m/s}$, $9.5\,\text{m/s}$, $8.5\,\text{m/s}$, $9.0\,\text{m/s}$, $8.0\,\text{m/s}$, $8.5\,\text{m/s}$, and $9.2\,\text{m/s}$, respectively.
\end{itemize}

The pertinent parameters of the nonlinear EV are shown in Table~\ref{table:Parameter_Settings}. The initial state vector and control input of the EV are set as $x_0 = [0, -6,0, 15, 0]^T$ and $u_0=[0,0]^T$, respectively. The following parameters are used: $N_c = 10$, $N=50$, $\textbf{Q}_m = diag(0,10^3,0,10^5,0)$, $w_1 =w_2 =w_3 = 5\times e^{-\frac{t}{50}}$, $ \textbf{Q}_T = diag(0,10^{9},10^{9}, 0,10^6)$, $\textbf{R} = diag(2\times10^4, 1\times10^6)$, $c = 8$, $\eta = 1$, $\varepsilon=10^{-5}$, $\gamma=50$, $\gamma_g = \gamma_c = \gamma_l=40$, $a = 3\,\text{m}$, $b=2\,\text{m}$, $p_{y,min} = -10.5\,\text{m}$, $p_{y,max}= -1.5\,\text{m}$, $M = 3$,  and $\textbf{w} = [2500, 150, 100,  100]^T$.

\begin{figure}[tp]
    \centering \hspace{-2.5mm}\includegraphics[scale=0.242]{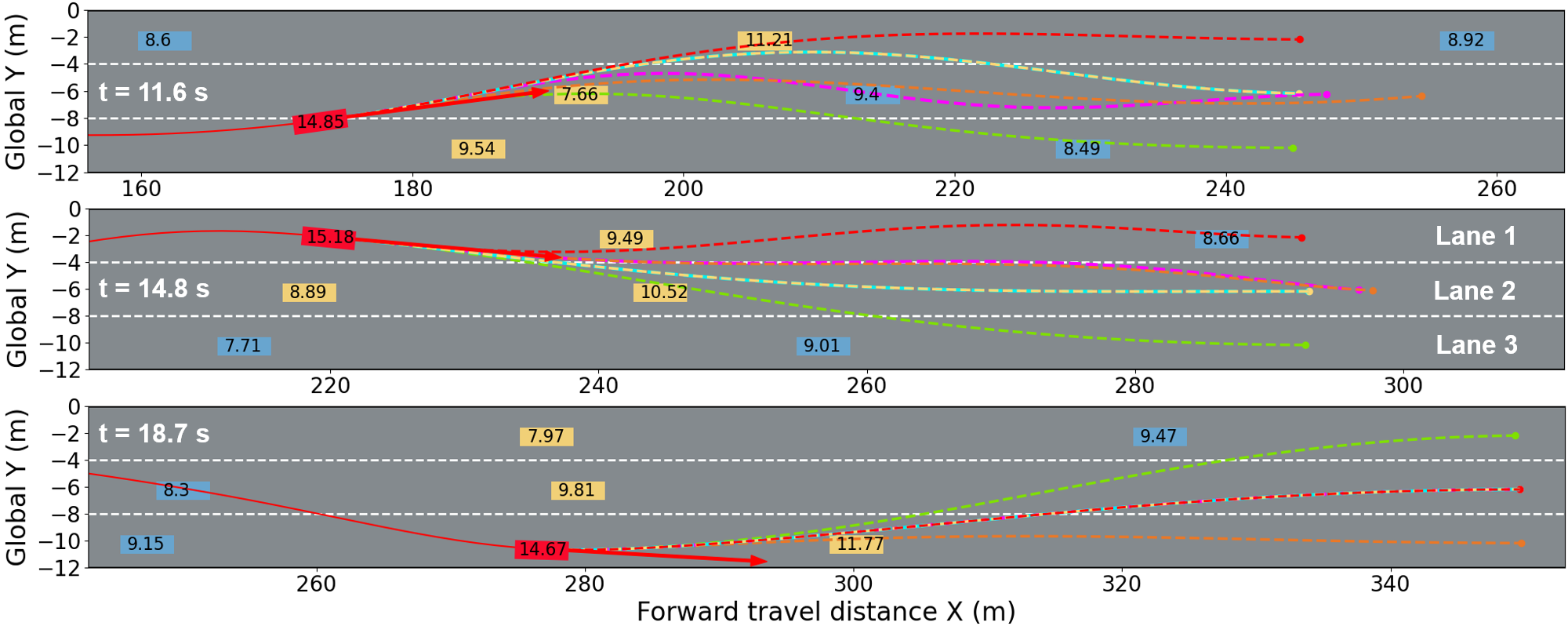}	
    \vspace{-1mm}
    \caption{\small{Snapshots of the EV's trajectory with the optimization horizon $T = 5\,\text{s}$ in dense and congested traffic. Red, yellow, and blue rectangles denote the EV, perceived, and unperceived SVs, respectively. Each dashed line represents a lane-oriented trajectory, and the red ones denote  the selected target trajectory. The red arrow denotes the current velocity vector of the AV. The number in each rectangle denotes the current velocity of each vehicle. Simulation video accessible at \protect\url{https://youtu.be/86oQ83jVPYg.}}}.
    \vspace{-6mm}
    \label{fig:tree6_snapots}
\end{figure}

 \begin{table}[tp]
    \centering
    \scriptsize
    \caption{Parameters of Vehicle Model}   \vspace{-1.5mm}
    \label{table:Parameter_Settings}
    \begin{tabular}{c c c c}
    \hline
    \hline 
    $v_{max}$  & $24\,\text{m/s} $  &$v_{min}$  & $0\,\text{m/s} $ \\ 
    $ \theta_{max}$  & $0.227\,\text{rad}$ & $\theta_{min}$ & $-0.227\,\text{rad}$\\
    $ \omega_{max}$  & $5\,\text{rad/s}$ & $\omega_{max}$ & $-5\,\text{rad/s}$\\
    $a_{max}$ & $3\,\text{m/s}^2$ & $a_{min}$  & $-1.5\,\text{m/s}^2$\\
    $\dot{\omega}_{max}$  &$ 2\,\text{rad/s}^2 $& $\dot{\omega}_{min}$  & $-2\,\text{rad/s}^2$\\
    \hline
    \hline
    \end{tabular}  \vspace{-2mm}
    \end{table}
\small
\begin{figure}[tp]
    \centering \hspace{-2.5mm}\includegraphics[scale=0.22]{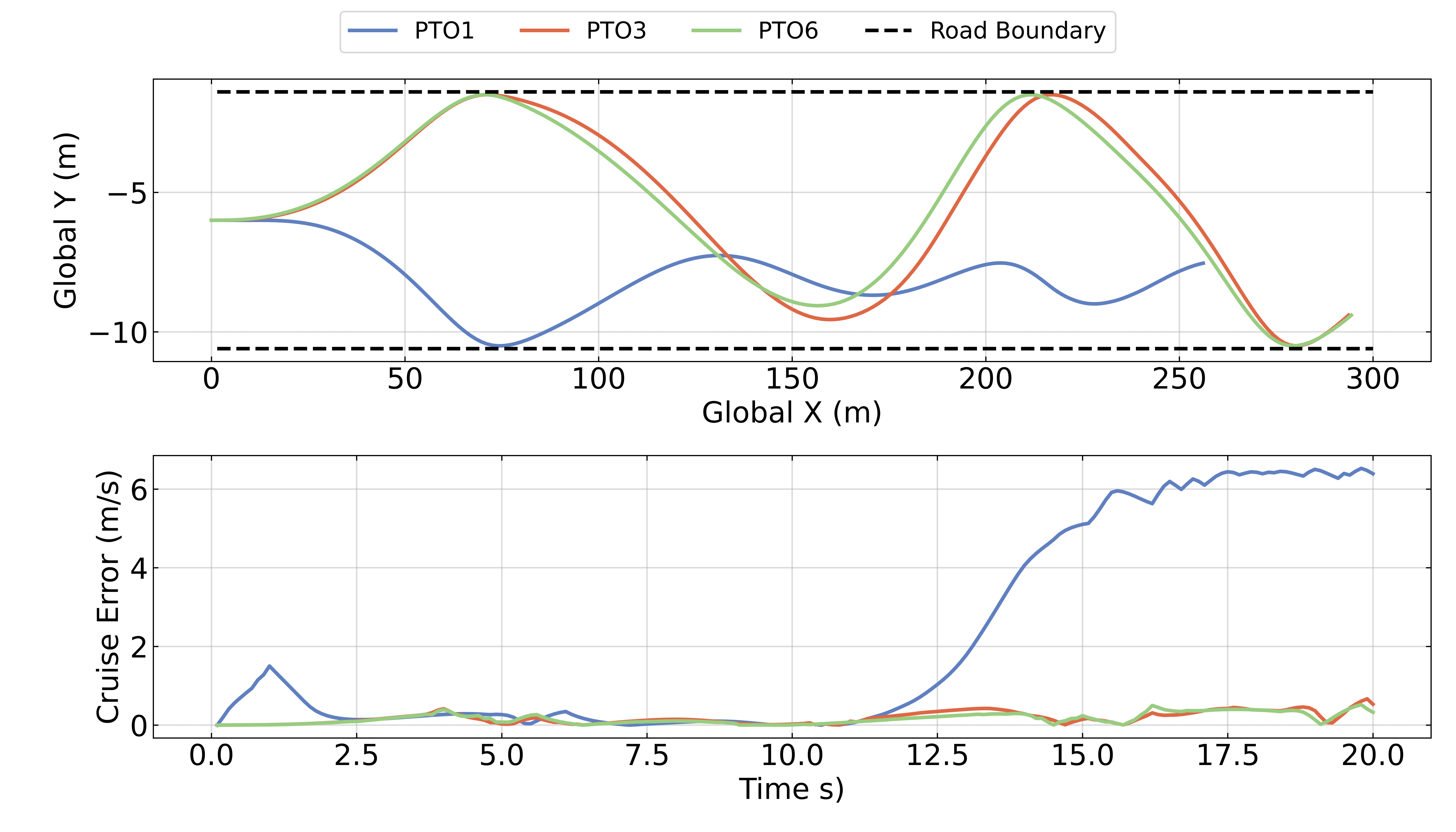}	
    \vspace{-2mm}
    \caption{\small{Cruise performance of the EV with a target cruise speed $15\,\text{m/s}$ and optimization horizon $T = 5\,\text{s}$ during $20\,\text{s}$ simulation.}}
    \vspace{-4mm}
    \label{fig:trajs}
\end{figure}
\begin{table*}[t]
    \centering
    \scriptsize
    \caption{Driving Performance Comparison}
    \label{tab:table_results}    
    \begin{tabular}[c]{|c | *{2}{c} | *{2}{c} | *{3}{c} | *{1}{c} |}
        %% HEADER
        \hline
        \multirow{2}{*}{\textbf{Algorithm}} & 
        \multicolumn{2}{c|}{\textbf{Cruise error}} &
        \multicolumn{2}{c|}{\textbf{Safety}} &
        \multicolumn{3}{c|}{\textbf{Efficiency}} &
        \multicolumn{1}{c|}{\textbf{Consistency}} \\
        & \mobility & \safety & \efficiency & \consistency \\
        \hline
        %% ENTRIES
        \multirow{1}{*}{\textbf{PTO1}} 
         & 2.136& 6.527 & 1.635 & $\mathbf{100}$  & $\mathbf{0.116}$ & $\mathbf{23.096}$ & 256.159 & - \\
        \hline
        \multirow{1}{*}{\textbf{PTO3}}
         & 0.182 & 0.668 & 2.760 & $\mathbf{100}$ & 0.148 &  24.553 & 293.705 & $\mathbf{100}$  \\
        \hline
        \multirow{1}{*}{\textbf{PTO6}}
         & $\mathbf{0.166}$ & $\mathbf{0.5184}$ &  $\mathbf{3.195}$ & $\mathbf{100}$& 0.150 &  28.690  & $\mathbf{294.407}$  & $\mathbf{100}$  \\
        %% END
        \hline
    \end{tabular}    \vspace{-5mm}
\end{table*}
\normalsize
% \subsection{Algorithms}\label{sec:pol_list}
We evaluate the following set of Algorithms:  
\begin{itemize}
    \item \textbf{PTO1}: This is an ablation study of our approach, achieved by solving (\ref{eq:obj})-(\ref{eq:domain}) with a single optimized trajectory that is consistent with the center lane.
    \item \textbf{PTO3}: Our proposed method, obtained by solving (\ref{eq:obj})-(\ref{eq:domain}) and (\ref{eq:optimal_behavior}), which optimizes over three nominal trajectories. Each trajectory corresponds to a unique lane in the three-lane road.
    \item \textbf{PTO6}: Similar to \textbf{PTO3}, it optimizes over six nominal trajectories. Four correspond to the center lane, and the remaining two correspond to the two outermost lanes.  This is illustrated in Fig.~\ref{fig:tree6_snapots}.
\end{itemize}

 \subsection{Results} 
    This subsection evaluates the performance of the three algorithms in terms of task accuracy, safety, driving efficiency, and driving stability, as shown in Table~\ref{tab:table_results}. The minimum barrier function value $\mathcal{S}_{min}$ of all three algorithms is positive with respect to the three nearest vehicles, ensuring safe interaction between the EV with uncertain SVs. Furthermore, PTO6 exhibits a significantly larger safety barrier value than PTO1 ($3.195 \,\text{m}$ versus $1.365 \,\text{m}$), with all candidate optimized trajectories satisfying the safety requirements with $\mathcal{P}_{safe} = 100\,\text{\%}$. These findings demonstrate that the spatiotemporal safety module enables the EV to safely interact with SVs.
 
   Figure~\ref{fig:trajs} depicts the evolution of cruise errors, providing an intuitive understanding of the cruise performance. Notably, the cruise error of PTO1 increases rapidly after $11\,\text{s}$, which is caused by the congested scenario in front of the EV. In contrast, PTO6 and PTO3 can enable the EV to escape from the congested driving scenario through multiple lane searching. To further elucidate the driving process, Fig.~\ref{fig:tree6_snapots} presents the trajectories and velocities of the EV during its escape from the congested driving scenario based on PTO6. As a result, PTO6 achieves the minimum deviation $e_{max}$ from the desired cruise speed $v_g$ among the three algorithms and reduces the mean absolute cruise error ${e}_{mean}$ by $92.223\%$ and $8.791\%$ compared to PTO3 and PTO1, respectively. These results highlight the advantage of using multiple parallel trajectory optimization methods in handling the uncertain behavior of SVs, which helps to reduce the impact of trajectory prediction errors resulting from uncertain SVs, leading to significant improvements in tracking accuracy.

\begin{figure}[tbp]
    \centering
\includegraphics[width=0.9\columnwidth]{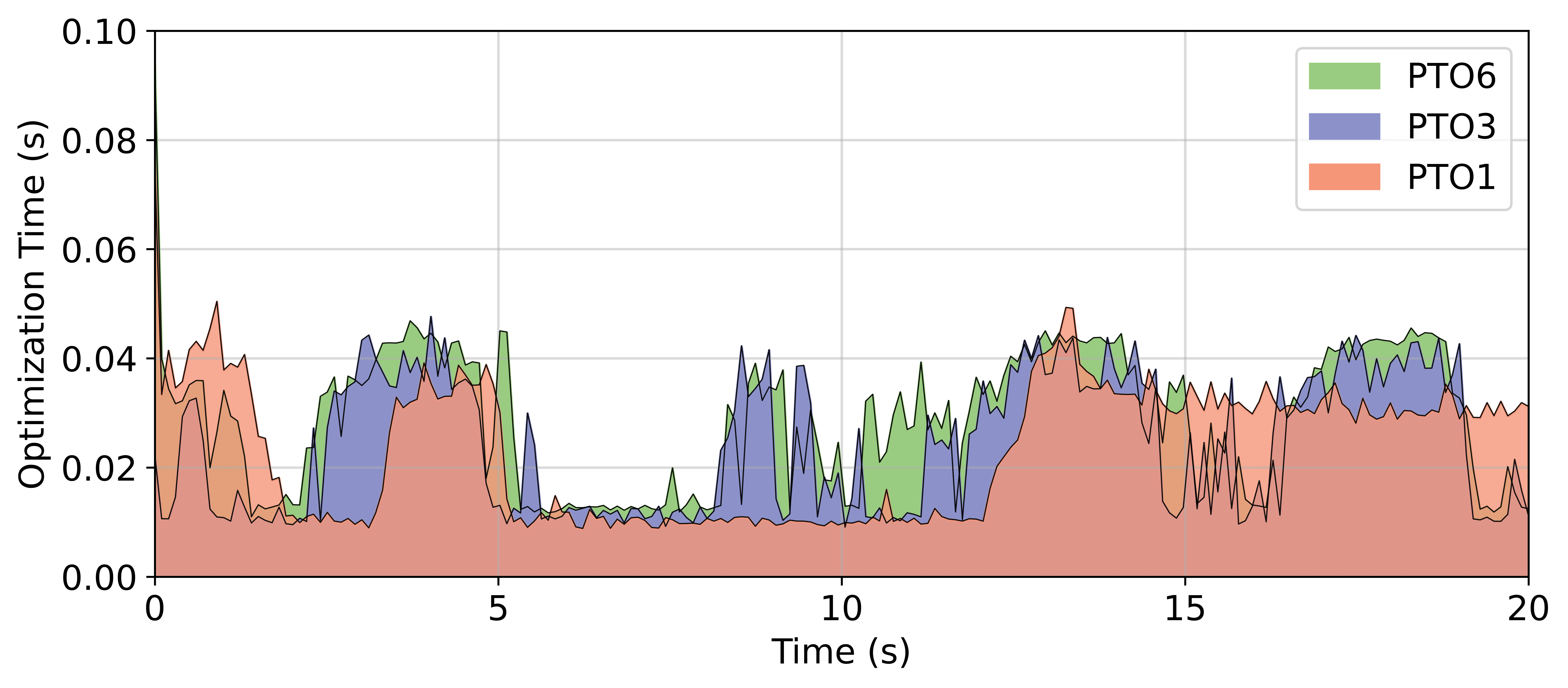}    \vspace{-1mm}
    \caption{\small{The evolution of optimization time of three algorithms with planning horizon $T = 5\,\text{s}$.}}
    \label{fig:optimization_time}        \vspace{-6mm}
\end{figure}

  In terms of driving efficiency, all three algorithms have an average optimization time $\mathcal{T}_{solve}$ less than $ 100\,\text{ms}$, and the specific optimization time evolution can be observed in Fig.~\ref{fig:optimization_time}. These findings demonstrate that these algorithms facilitate real-time replanning of the EV in this congested traffic. Regarding travel efficiency, PTO3 and PTO6 exhibit a notably longer travel distance $\mathcal{L}_{long}$ than PTO1 (294.407$\,\text{m}$ and 293.705$\,\text{m}$ versus 265.159$\,\text{m}$). These results demonstrate that our framework significantly improves driving efficiency while maintaining a low average acceleration $\mathcal{A}_{mean}$ and short solving time $\mathcal{T}_{solve}$ through leveraging multiple searching trajectories.
% \begin{figure}[tp]
% \hspace{-5mm} 
% \begin{subfigure}[b]{0.25\textwidth}
%     \centering
%     \includegraphics[scale=0.38]{figs/tree_3_lane.png}
%     \caption{PTO3}
%     \label{fig:consistency_3}
% \end{subfigure}
% \hspace{-6mm} % adjust the length as needed
% \begin{subfigure}[b]{0.25\textwidth}
%     \centering
%     \includegraphics[scale=0.38]{figs/tree_6_lane.png}
%     \caption{PTO6}
%     \label{fig:consistency_6}
% \end{subfigure}
% \caption{The evolution of the EV's target lane.}\vspace{-6mm}
% \label{fig:consistency}
% \end{figure} 
\begin{figure}[tp]
    \centering
        \subfigure[PTO3]{
            \label{fig:consistency_3}
        \includegraphics[scale=0.38]{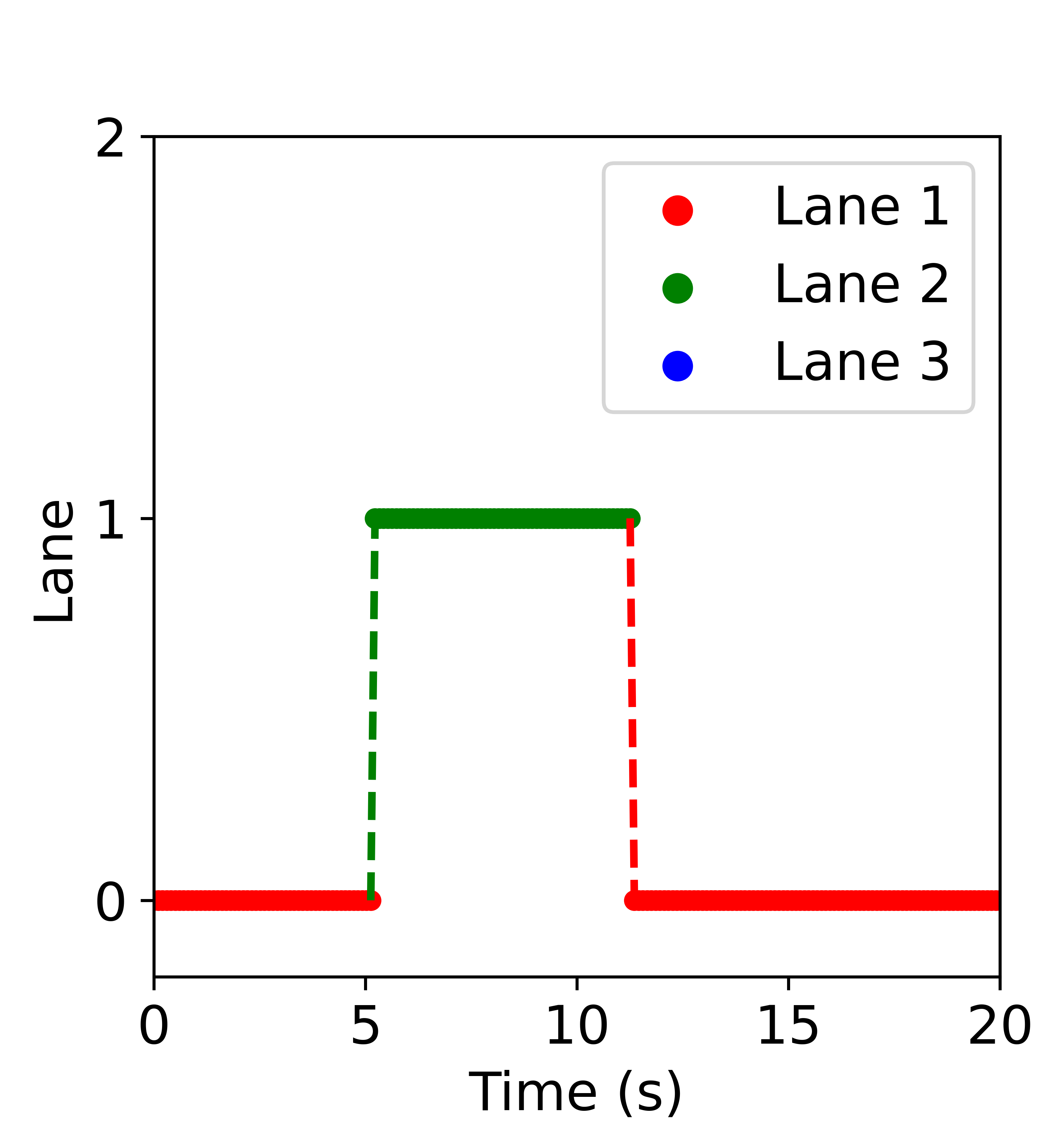}}\hspace{-2mm} 
        \subfigure[PTO6]{
            \label{fig:consistency_6}
        \includegraphics[scale=0.38]{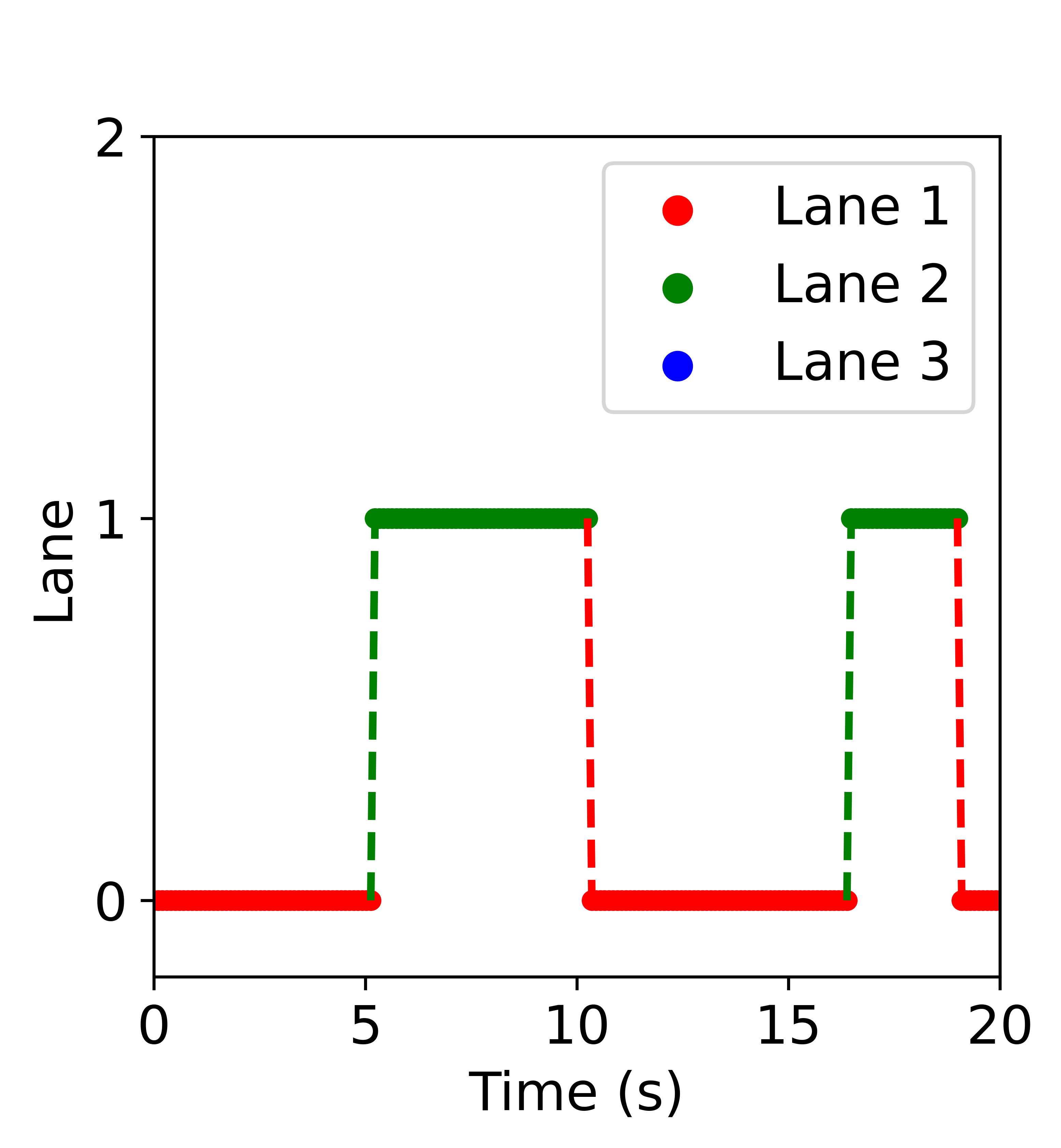}}\hspace{-2mm}
    \vspace{-1mm}
    \caption{\small{The evolution of the EV's target lane.}}	
    \label{fig:consistency}
    \vspace{-5mm}
\end{figure}	  
 With regard to driving stability, Fig.~\ref{fig:consistency} illustrates the evolution of the selected target lane during driving, revealing that there is no sudden lane change during each phase, resulting in good driving consistency. This finding is further supported by the consistency percentage of decision-making maneuvers $\mathcal{P}_{\text{LC}} = 100\%$, which indicates that there is no abrupt lane change during the driving task, further confirming the EV's good driving consistency.

\section{Conclusion}
\label{sec:conclusion}
In this paper, we presented a parallel trajectory optimization method for fast replanning for the EV  to achieve high travel efficiency in dense and congested traffic. Our approach efficiently generates feasible and safe candidate trajectories, while maintaining high computational efficiency based on the multiple-shooting and multi-threading techniques. Our simulation results demonstrated the effectiveness of our framework in improving safety, reducing travel time, and maintaining driving consistency for the EV in dense and congested traffic flow. In future work, we plan to extend our framework to address multi-agent navigation problems.

% \input{Sections/conclusion}
%\section*{ACKNOWLEDGMENT}

% \newpage
\bibliographystyle{IEEEtran}
\bibliography{references.bib}

\begin{thebibliography}{10}
\providecommand{\url}[1]{#1}
\csname url@rmstyle\endcsname
\providecommand{\newblock}{\relax}
\providecommand{\bibinfo}[2]{#2}
\providecommand\BIBentrySTDinterwordspacing{\spaceskip=0pt\relax}
\providecommand\BIBentryALTinterwordstretchfactor{4}
\providecommand\BIBentryALTinterwordspacing{\spaceskip=\fontdimen2\font plus
\BIBentryALTinterwordstretchfactor\fontdimen3\font minus \fontdimen4\font\relax}
\providecommand\BIBforeignlanguage[2]{{%
\expandafter\ifx\csname l@#1\endcsname\relax
\typeout{** WARNING: IEEEtran.bst: No hyphenation pattern has been}%
\typeout{** loaded for the language `#1'. Using the pattern for}%
\typeout{** the default language instead.}%
\else
\language=\csname l@#1\endcsname
\fi
#2}}

\bibitem{schrank2019urban}
D.~Schrank, B.~Eisele, T.~Lomax, \emph{et~al.}, ``Urban mobility report 2019,'' Texas A\&M Transportation Institute, College Station, Technical Report, 2019.

\bibitem{narayanan2020shared}
S.~Narayanan, E.~Chaniotakis, and C.~Antoniou, ``Shared autonomous vehicle services: A comprehensive review,'' \emph{Transportation Research Part C: Emerging Technologies}, vol. 111, pp. 255--293, 2020.

\bibitem{liu2021role}
T.~Liu \emph{et~al.}, ``The role of the hercules autonomous vehicle during the {COVID-19} pandemic: An autonomous logistic vehicle for contactless goods transportation,'' \emph{IEEE Robotics \& Automation Magazine}, vol.~28, no.~1, pp. 48--58, 2021.

\bibitem{tian2021trajectory}
F.~Tian, R.~Zhou, Z.~Li, L.~Li, Y.~Gao, D.~Cao, and L.~Chen, ``Trajectory planning for autonomous mining trucks considering terrain constraints,'' \emph{IEEE Transactions on Intelligent Vehicles}, vol.~6, no.~4, pp. 772--786, 2021.

\bibitem{shalev2017formal}
S.~Shalev-Shwartz, S.~Shammah, and A.~Shashua, ``On a formal model of safe and scalable self-driving cars,'' \emph{arXiv preprint arXiv:1708.06374}, 2017.

\bibitem{leung2020infusing}
K.~Leung, E.~Schmerling, M.~Zhang, M.~Chen, J.~Talbot, J.~C. Gerdes, and M.~Pavone, ``On infusing reachability-based safety assurance within planning frameworks for human--robot vehicle interactions,'' \emph{The International Journal of Robotics Research}, vol.~39, no. 10-11, pp. 1326--1345, 2020.

\bibitem{werling2012optimal}
M.~Werling, S.~Kammel, J.~Ziegler, and L.~Gr{\"o}ll, ``Optimal trajectories for time-critical street scenarios using discretized terminal manifolds,'' \emph{The International Journal of Robotics Research}, vol.~31, no.~3, pp. 346--359, 2012.

\bibitem{zhang2020optimal}
Y.~Zhang, H.~Sun, J.~Zhou, J.~Pan, J.~Hu, and J.~Miao, ``Optimal vehicle path planning using quadratic optimization for {B}aidu {A}pollo open platform,'' in \emph{IEEE Intelligent Vehicles Symposium}.\hskip 1em plus 0.5em minus 0.4em\relax IEEE, 2020, pp. 978--984.

\bibitem{sharath2020enhanced}
M.~Sharath and N.~R. Velaga, ``Enhanced intelligent driver model for two-dimensional motion planning in mixed traffic,'' \emph{Transportation Research Part C: Emerging Technologies}, vol. 120, p. 102780, 2020.

\bibitem{miller2018efficient}
C.~Miller, C.~Pek, and M.~Althoff, ``Efficient mixed-integer programming for longitudinal and lateral motion planning of autonomous vehicles,'' in \emph{IEEE Intelligent Vehicles Symposium}.\hskip 1em plus 0.5em minus 0.4em\relax IEEE, 2018, pp. 1954--1961.

\bibitem{liu2017speed}
C.~Liu, W.~Zhan, and M.~Tomizuka, ``Speed profile planning in dynamic environments via temporal optimization,'' in \emph{IEEE Intelligent Vehicles Symposium}.\hskip 1em plus 0.5em minus 0.4em\relax IEEE, 2017, pp. 154--159.

\bibitem{fan2018baidu}
H.~Fan, F.~Zhu, C.~Liu, L.~Zhang, L.~Zhuang, D.~Li, W.~Zhu, J.~Hu, H.~Li, and Q.~Kong, ``Baidu {A}pollo {EM} motion planner,'' \emph{arXiv preprint arXiv:1807.08048}, 2018.

\bibitem{jian2020multi}
Z.~Jian, S.~Chen, S.~Zhang, Y.~Chen, and N.~Zheng, ``Multi-model-based local path planning methodology for autonomous driving: An integrated framework,'' \emph{IEEE Transactions on Intelligent Transportation Systems}, vol.~23, no.~5, pp. 4187--4200, 2020.

\bibitem{xu2021autonomous}
W.~Xu, Q.~Wang, and J.~M. Dolan, ``Autonomous vehicle motion planning via recurrent spline optimization,'' in \emph{IEEE International Conference on Robotics and Automation}.\hskip 1em plus 0.5em minus 0.4em\relax IEEE, 2021, pp. 7730--7736.

\bibitem{Gamann2019TowardsSO}
B.~Gassmann, F.~Oboril, C.~Buerkle, S.~Liu, S.~Yan, M.~S. Elli, I.~Alvarez, N.~Aerrabotu, S.~Jaber, P.~van Beek, D.~Iyer, and J.~Weast, ``Towards standardization of av safety: C++ library for responsibility sensitive safety,'' in \emph{IEEE Intelligent Vehicles Symposium}, 2019, pp. 2265--2271.

\bibitem{mayne2014model}
D.~Q. Mayne, ``Model predictive control: Recent developments and future promise,'' \emph{Automatica}, vol.~50, no.~12, pp. 2967--2986, 2014.

\bibitem{zeng2021safety}
J.~Zeng, B.~Zhang, and K.~Sreenath, ``Safety-critical model predictive control with discrete-time control barrier function,'' in \emph{American Control Conference}.\hskip 1em plus 0.5em minus 0.4em\relax IEEE, 2021, pp. 3882--3889.

\bibitem{zheng2022safe}
L.~Zheng, R.~Yang, Z.~Wu, J.~Pan, and H.~Cheng, ``Safe learning-based gradient-free model predictive control based on cross-entropy method,'' \emph{Engineering Applications of Artificial Intelligence}, vol. 110, p. 104731, 2022.

\bibitem{yin2022trajectory}
J.~Yin, Z.~Zhang, E.~Theodorou, and P.~Tsiotras, ``Trajectory distribution control for model predictive path integral control using covariance steering,'' in \emph{IEEE International Conference on Robotics and Automation}, 2022, pp. 1478--1484.

\bibitem{wurts2018collision}
J.~Wurts, J.~L. Stein, and T.~Ersal, ``Collision imminent steering using nonlinear model predictive control,'' in \emph{American Control Conference}.\hskip 1em plus 0.5em minus 0.4em\relax IEEE, 2018, pp. 4772--4777.

\bibitem{ames2019control}
A.~D. Ames, S.~Coogan, M.~Egerstedt, G.~Notomista, K.~Sreenath, and P.~Tabuada, ``Control barrier functions: Theory and applications,'' in \emph{European Control Conference}.\hskip 1em plus 0.5em minus 0.4em\relax IEEE, 2019, pp. 3420--3431.

\bibitem{sadat2019jointly}
A.~Sadat, M.~Ren, A.~Pokrovsky, Y.-C. Lin, E.~Yumer, and R.~Urtasun, ``Jointly learnable behavior and trajectory planning for self-driving vehicles,'' in \emph{IEEE/RSJ International Conference on Intelligent Robots and Systems}.\hskip 1em plus 0.5em minus 0.4em\relax IEEE, 2019, pp. 3949--3956.

\bibitem{ma2022local}
J.~Ma, Z.~Cheng, X.~Zhang, Z.~Lin, F.~L. Lewis, and T.~H. Lee, ``Local learning enabled iterative linear quadratic regulator for constrained trajectory planning,'' \emph{IEEE Transactions on Neural Networks and Learning Systems}, 2022.

\bibitem{boyd2011distributed}
S.~Boyd, N.~Parikh, E.~Chu, B.~Peleato, J.~Eckstein, \emph{et~al.}, ``Distributed optimization and statistical learning via the alternating direction method of multipliers,'' \emph{Foundations and Trends{\textregistered} in Machine learning}, vol.~3, no.~1, pp. 1--122, 2011.

\bibitem{ma2022alternating}
J.~Ma, Z.~Cheng, X.~Zhang, M.~Tomizuka, and T.~H. Lee, ``Alternating direction method of multipliers for constrained iterative {LQR} in autonomous driving,'' \emph{IEEE Transactions on Intelligent Transportation Systems}, vol.~23, no.~12, pp. 23\,031--23\,042, 2022.

\bibitem{han2023rda}
R.~Han, S.~Wang, S.~Wang, Z.~Zhang, Q.~Zhang, Y.~C. Eldar, Q.~Hao, and J.~Pan, ``{RDA}: An accelerated collision free motion planner for autonomous navigation in cluttered environments,'' \emph{IEEE Robotics and Automation Letters}, vol.~8, no.~3, pp. 1715--1722, 2023.

\bibitem{adajania2022multi}
V.~K. Adajania, A.~Sharma, A.~Gupta, H.~Masnavi, K.~M. Krishna, and A.~K. Singh, ``Multi-modal model predictive control through batch non-holonomic trajectory optimization: Application to highway driving,'' \emph{IEEE Robotics and Automation Letters}, vol.~7, no.~2, pp. 4220--4227, 2022.

\bibitem{chen2017constrained}
J.~Chen, W.~Zhan, and M.~Tomizuka, ``Constrained iterative {LQR} for on-road autonomous driving motion planning,'' in \emph{IEEE International Conference on Intelligent Transportation Systems}, 2017, pp. 1--7.

\bibitem{zheng2023STRHC}
L.~Zheng, R.~Yang, Z.~Peng, M.~Y. Wang, and J.~Ma, ``Spatiotemporal receding horizon control with proactive interaction towards safe and efficient autonomous driving in dense traffic,'' \emph{arXiv preprint arXiv:2308.05929}, 2023.

\bibitem{de1998stabilizing}
G.~De~Nicolao, L.~Magni, and R.~Scattolini, ``Stabilizing receding-horizon control of nonlinear time-varying systems,'' \emph{IEEE Transactions on Automatic Control}, vol.~43, no.~7, pp. 1030--1036, 1998.

\bibitem{fontes2001general}
F.~A. Fontes, ``A general framework to design stabilizing nonlinear model predictive controllers,'' \emph{Systems \& Control Letters}, vol.~42, no.~2, pp. 127--143, 2001.

\bibitem{betts2010practical}
J.~T. Betts, \emph{{\emph{Practical Methods for Optimal Control and Estimation Using Nonlinear Programming}}, 2nd ed}.\hskip 1em plus 0.5em minus 0.4em\relax Philadelphia, PA: SIAM, 2010.

\bibitem{borrelli2017predictive}
F.~Borrelli, A.~Bemporad, and M.~Morari, \emph{{\emph{Predictive Control for Linear and Hybrid Systems}}}.\hskip 1em plus 0.5em minus 0.4em\relax New York, NY, USA: Cambridge University Press, 2017.

\bibitem{mastalli2022feasibility}
C.~Mastalli, W.~Merkt, J.~Marti-Saumell, H.~Ferrolho, J.~Sol{\`a}, N.~Mansard, and S.~Vijayakumar, ``A feasibility-driven approach to control-limited {DDP},'' \emph{Autonomous Robots}, vol.~46, no.~8, pp. 985--1005, 2022.

\bibitem{gratton2007approximate}
S.~Gratton, A.~S. Lawless, and N.~K. Nichols, ``Approximate {Gauss--Newton} methods for nonlinear least squares problems,'' \emph{SIAM Journal on Optimization}, vol.~18, no.~1, pp. 106--132, 2007.

\bibitem{Houska2011a}
B.~Houska, H.~Ferreau, and M.~Diehl, ``{ACADO} {T}oolkit -- {A}n {O}pen {S}ource {F}ramework for {A}utomatic {C}ontrol and {D}ynamic {O}ptimization,'' \emph{Optimal Control Applications and Methods}, vol.~32, no.~3, pp. 298--312, 2011.

\end{thebibliography}

\end{document}